# Image Quality, Uniformity and Computation Improvement of Compressive Light Field Displays with U-Net

*Chen Gao\*, Haifeng Li\*\*, Xu Liu\*\*, Xiaodi Tan\*,*
\*College of Photonic and Electronic Engineering, Fujian Normal University, Fuzhou, China
\*\*College of Optical Science and Engineering, Zhejiang University, Hangzhou, China

**Abstract**
*We apply the U-Net model for compressive light field synthesis. Compared to methods based on stacked CNN and iterative algorithms, this method offers better image quality, uniformity and less computation.*

## 1. Introduction

Three-dimensional (3D) display technology is the entrance to the realistic-feeling metaverse for tabletop, portable and near-eye electronic devices. True 3D displays are mainly divided into light field displays and holographic displays. Compressive light field displays utilize the scattering characteristic of display panels and the correlation between viewpoint images of the 3D scene [1]. The compressive light field display is a candidate for portable 3D displays owing to its compact structure, moderate viewing angle and high spatial resolution. However, computational resources of portable electronic devices are restricted to satisfy their duration demand. Iterative algorithms to solve the compressive light field display images have the problem of heavy computation [2], which prevents the compressive light field display from being a practical solution for portable dynamic 3D displays. With the development of artificial intelligence (AI) technology, image synthesis algorithms based on deep learning are gradually applied to 3D displays [3]. Deep neural networks can be trained to fit the iterative process. Moreover, fast display image synthesis could be realized with rapid forward propagation of artificial neural networks (ANN). Former researchers proposed a stacked CNN-based method for compressive light field synthesis [4]. However, the stacked convolution neural network (CNN)-based method suffers from convergence and over-fitting problems. Quan et al. [5] propose an image synthesis method for curved compressive light field displays based on Neural Radiance Fields (NeRF). But the NeRF-based approach learns the voxel information of a single scene rather than the structural information of the display system [6]. The network must be retrained once the display scene changes, so this method is unsuitable for dynamic display devices. Sun et al. [7] propose a depth-assisted calibration on learning-based factorization for compressive light field displays. Nevertheless, this method still needs patterns solved by the iterative algorithm as priori knowledge, and its calculation time is the sum of the time spent by the iterative algorithm and the network inference process. Therefore, applying this method for real-time compressive light field displays is difficult.

U-Net is initially used for image segmentation in computed tomography (CT) to handle slicing data and output the organ's cancer probability [8]. The skip connection added in the U-Net architecture significantly improves its convergence compared with the stacked CNN model. Light field data are pretty similar to slicing data in CT. Thus, this paper introduces U-Net as the network model for optimizing compressive light field display images for better convergence and generalization. Given a specific viewing angle, several augmented target light field datasets are generated as the training set of U-Net. After the U-Net converges, the trained U-Net synthesizes the display images that reconstruct the target light field for testing. The results of training and testing prove that, compared to the method based on stacked CNN and iterative algorithms, the proposed U-Net-based image synthesis method for compressive light field displays has the strength of higher reconstruction quality and fewer computing resources.

## 2. Principle

The image synthesis principle of compressive light field displays based on ANN is illustrated in Fig. 1. The training procedure of ANN is split into forward and backward propagation. The forward propagation includes two steps: firstly, input the target light field as training data into the ANN; then, simulate the reconstructed light field with perspective projection of the output images. The backward propagation updates ANN's parameters with loss function taking the reconstructed and target light fields as variables. Repeat the above procedure in each training epoch and batch until the ANN converges. Once the training is finished, the trained ANN can be used in inference procedures: with the target light field input as test data, display images are synthesized directly without iterations. The training and inference procedures with ANNs of different architectures are similar. Thus, the effect of training and inference lies in the option of network architecture and training hyper-parameters.

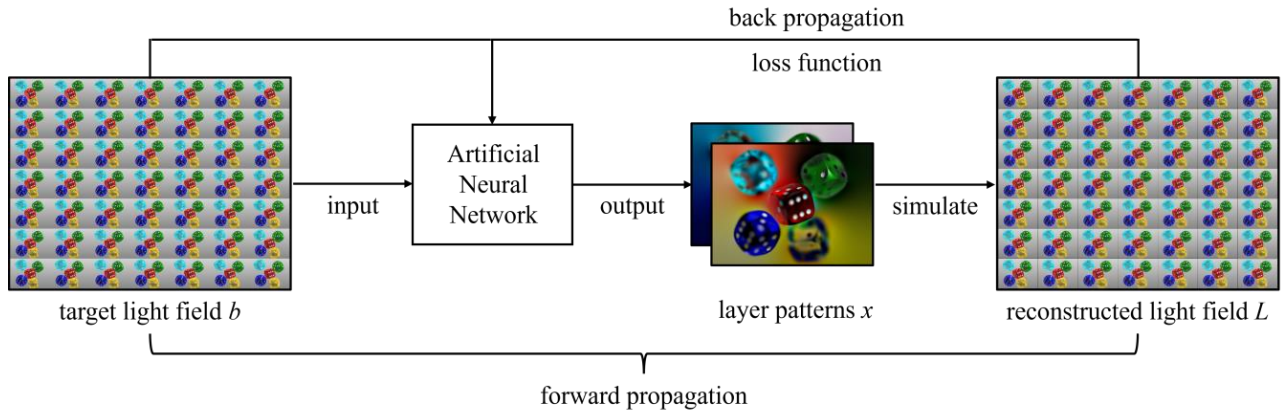

**Figure 1.** Principle of compressive light field synthesis based on artificial neural network.

## 3. U-Net architecture for compressive light field synthesis

Toshiaki et al. [4] propose a stacked CNN architecture for compressive light field synthesis. The network is composed of 19 identical modules stacked together. Every module comprises a 64-channel 2D convolution operation and a Rectified Linear Unit (ReLU). Input and output channels are matched with the viewpoint number of the target light field and display layer image number, respectively. Input, stacked CNN modules and output are connected by a ReLU function. However, the stacked CNN architecture has the problem of vanishing gradient. When error gradients are back-propagated into earlier layers, repeated multiplication makes the gradient disappear. As a result, the more modules there are in the network, its performance tends to saturate and decline rapidly. To solve the vanishing gradient problem of stacked CNN, researchers propose other network architectures for various task targets, and U-Net is one of them for image segmentation. There is a close relationship between compressive light field displays and CT imaging. U-Net is used to process slice data from CT scans to output the probability of an organ becoming cancerous based on the attenuation coefficient between adjacent slices. The probability of an organ becoming cancerous is represented by two double-channel probability maps. The formation of slice data in CT imaging is similar to light field data in 3D displays. If the input and output of the U-Net are changed to the target light field and multi-layer display images, respectively, then a U-Net can be used for compressive light field synthesis. The U-Net architecture for compressive light field synthesis is shown in Fig. 2. The basic operations of U-Net can be divided into convolution, pooling, and up-convolution. After the target light field passes through the input module, the channel becomes 64, and the channel after each convolution pooling operation will become twice as before. When the data reaches the bottom of the network, the channel is changed back to 64 through symmetric up-convolution and convolution operations, and finally a convolution operation with a kernel of 1×1 is used to match the output channels with the number of layered display images. The key to U-Net's ability to avoid the vanishing gradient problem is the skip-connection between each symmetrical module of the down and up paths, which copies the parameters of the down path directly to the corresponding up path. The addition of skip-connections is based on the following assumption: when processing image segmentation tasks, the segmentation results obtained should be similar to the original images at different levels. There are also similarities between the compressive light field display images and the target light field, which is the theoretical foundation for using U-Net to synthesize compressive light field displays.

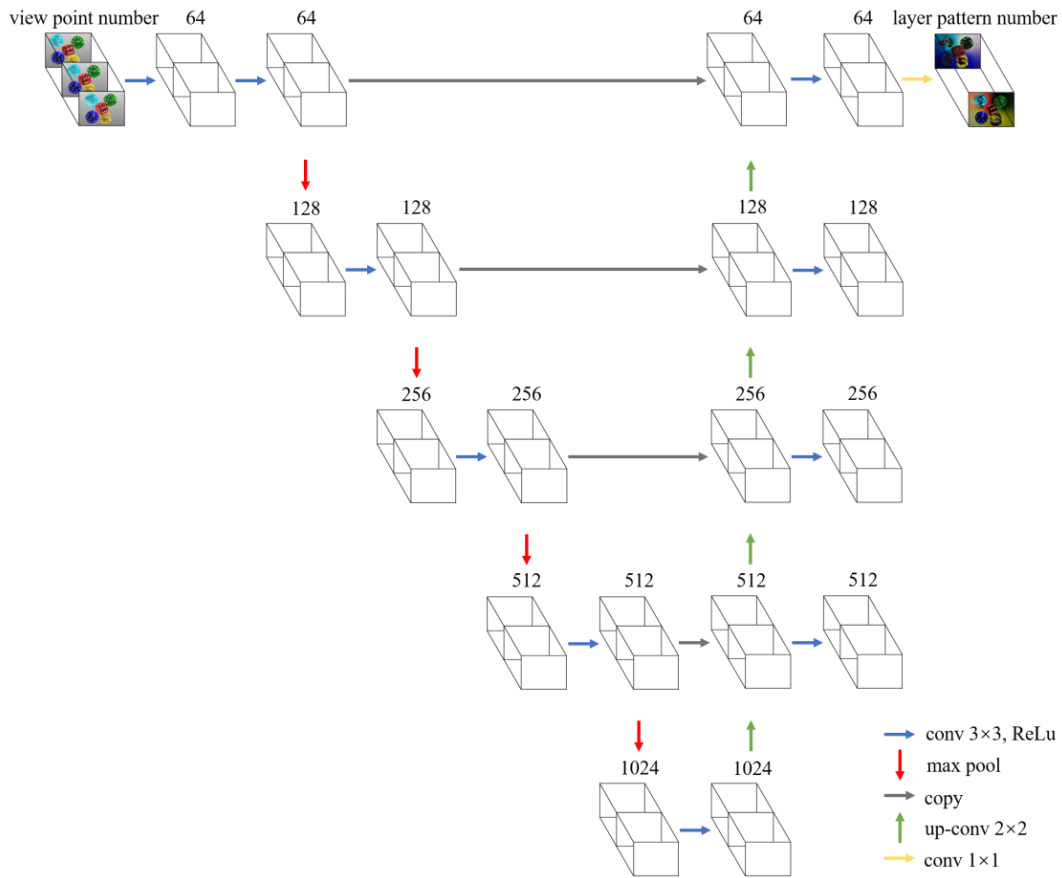

**Figure 2.** U-Net architecture for compressive light field synthesis.

Current AI technology is driven by prior knowledge and data. The prior knowledge lies in the design of ANN structures according to the characteristics of various tasks. When training ANN, the accuracy of training datasets can seriously affect the model's results because the network learns biases in training datasets, and the design of training datasets also depends on prior knowledge of the task. Images of the compressive light field display at different depths are synthesized from perspective projections of viewpoints. This process is a constant mapping relationship from angles to depths. Besides, the ray modulation (e.g., multiplicative or additive) between the display layers is also considered in synthesis. One training trick U-Net authors suggested is using data augmentation to increase the number of input images, thus eliminating the need for additional labeled data. This data augmentation is accomplished through "elastic deformation" that changes the object's shape in the image as if it were in a different position. However, this data augmentation method is unsuitable for the training task proposed in this paper because when the viewpoint images are changed by rotation, scaling and other transformations, the mapping relationship between angles and depths will also change. We adopt the data augmentation procedure proposed by Toshiaki et al. [4], i.e., augment the pixel intensity and reduce the pixel number of training datasets. Because we find that if the target light field with a large number of pixels is directly input into the network as training data, the output result will converge to unchanged images whose gray level is close to the scene background, which indicates that the network learns the background information of the scene. In addition, for objects whose textures change slowly, the boundaries between them and the background imply position changes of pixels.

To make a fair comparison with the stacked CNN-based approach proposed by Toshiaki et al. [4], we set other training hyperparameters except the network architecture to be consistent, including:

(1) The training datasets are the same.

(2) All activation functions are ReLU.

(3) All network parameters are initialized uniformly and randomly using Kaiming Initialization.

(4) The optimizer is Adam: the learning rate is set to a small value of $10^{-4}$ to avoid training results from oscillating, and other parameters are default values. After the first training batch, the Adam optimizer will automatically adjust the learning rate.

(5) There are enough training epochs to ensure network convergence, the number of training epochs is set to 100, and each training batch consists of 15 data.

(6) The loss function is the mean square error between the reconstructed and target light fields, and the regularization term is the pixel value of display images.

## 4. Result

A target light field (shown in Fig. 3) is input into the trained networks to test their generalization. The target light field's viewing angle and viewpoint number for testing are identical to training datasets. Here, we take the additive type of compressive light field display for demonstration. Peak signal-to-noise ratio (PSNR) between the reconstructed and target light fields is used as evaluation metrics. Fig.4 shows the training and test PSNRs of stacked CNN and U-Net varying with epochs. The training and test results show that U-Net can better fit and generalize compressive light field synthesis than stacked CNN. As shown in Fig.5, the network model with the best testing results is selected to synthesize display images and simulate the reconstructed light field, and compared with the images synthesized by the iterative algorithm of 100 iterations. The display images synthesized by stacked CNN and U-Net have apparent differences, but both conform to the regularization term of the pixel value range. But stacked CNN tends to allocate as many pixel values as possible to the first layer, and pixel value changes in the remaining layers gradually decrease. As Fig. 6 shows, U-Net's synthesis demonstrates better reconstruction quality and uniformity, comparable to the iterative reconstruction of about 50 iterations.

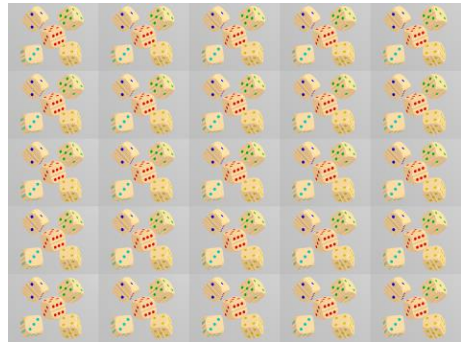

**Figure 3.** Target light field for testing.

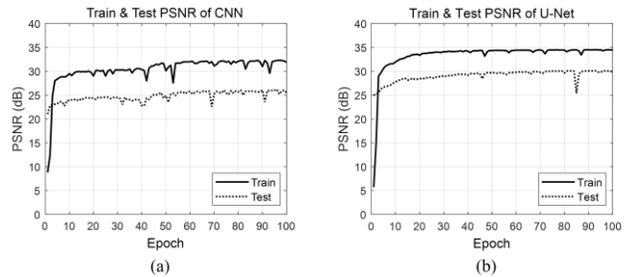

**Figure 4.** Training and testing results of networks for additive light field synthesis. (a) stacked CNN. (b) U-Net.

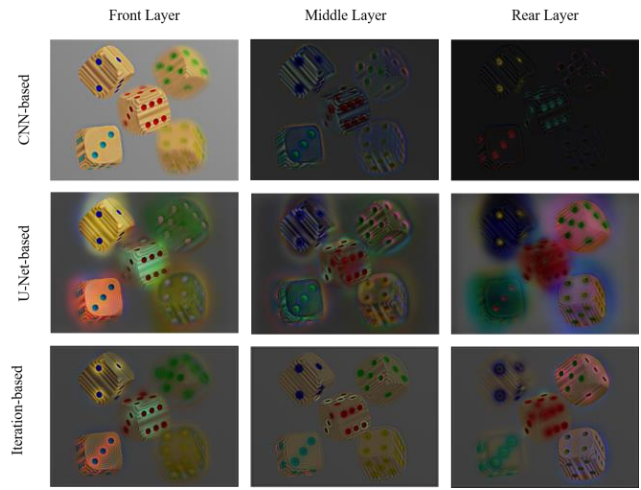

**Figure 5.** Additive light field display images synthesized by stacked CNN, U-Net and iterative algorithm.

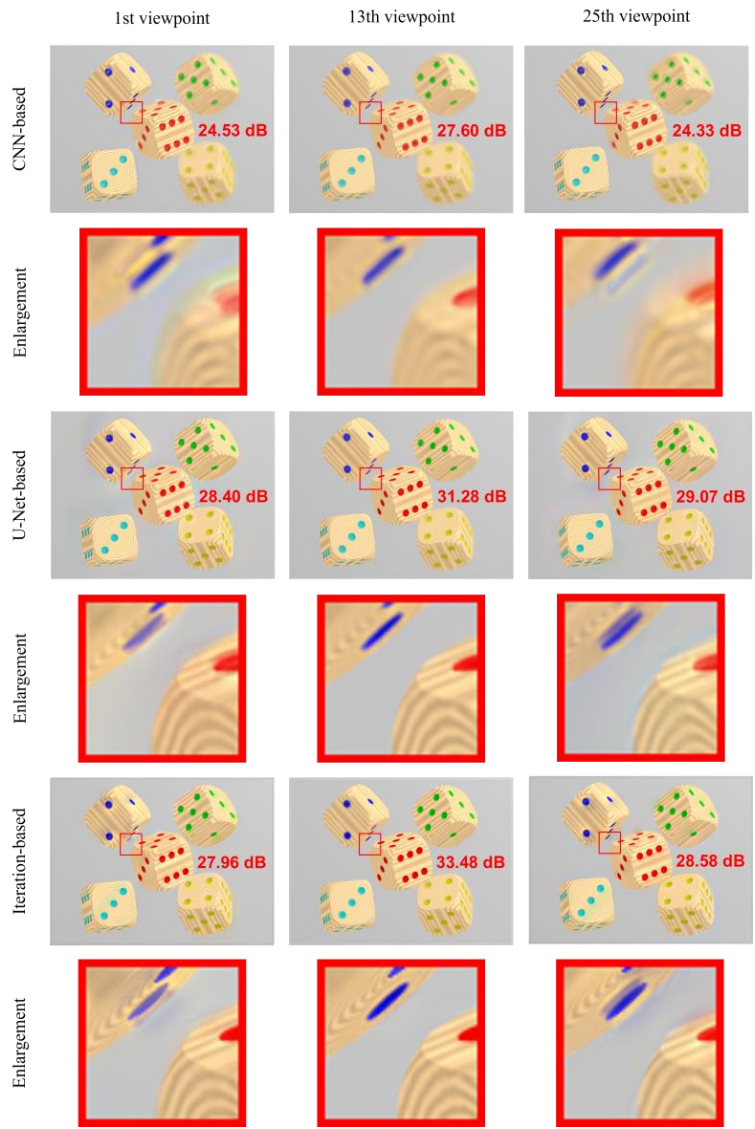

**Figure 6.** Simulated reconstruction of viewpoint images by the additive light field display.

## 5. Acknowledgements
This work is supported by the Natural Science Foundation of China (U22A2080).